%% file: TPAMI-2025-06-1711.R1_Makabe.tex
\documentclass[lettersize,journal]{IEEEtran}
\input{mlpreamble}

\usepackage[accsupp]{axessibility}  
\usepackage{hyperref}
\usepackage[whole]{bxcjkjatype}
\usepackage{multirow}
\usepackage{cases}
\usepackage{array}
\usepackage{wrapfig}
\usepackage{slashbox}
\usepackage{overpic}
\usepackage{subcaption}

\begin{document}

\title{\journaltitle} 
\author{
Lilika~Makabe,
Kohei Ashida,
Hiroaki~Santo,~\IEEEmembership{Member,~IEEE,}
Fumio~Okura,~\IEEEmembership{Member,~IEEE,} 
and Yasuyuki~Matsushita,~\IEEEmembership{Fellow,~IEEE}
\IEEEcompsocitemizethanks{\IEEEcompsocthanksitem L. Makabe, K.Ashida, H. Santo, F. Okura, and Y. Matsushita are with Graduate School of Information Science and Technology, The University of Osaka, Japan. 
\protect\\
E-mail: \{makabe.lilika, santo.hiroaki, okura, yasumat\}@ist.osaka-u.ac.jp 
}
}


\markboth{Journal of \LaTeX\ Class Files,~2026}%
{Makabe \MakeLowercase{\textit{et al.}}: DP-SfM: Dual-Pixel Structure-from-Motion without Scale Ambiguity}




\maketitle

\input{sec/0_abstract}    
\input{sec/01_intro}

\input{sec/02_related}
\input{sec/03_method}
\input{sec/04_experiments}

\input{sec/05_conclusion}


%
%
\bibliographystyle{IEEEtran}
\bibliography{main}
\end{document}

%% file: mlpreamble.tex
\usepackage{booktabs}
\usepackage{xcolor}
\usepackage{color, colortbl}
\usepackage{amsmath, amssymb, amsfonts}
\usepackage{graphicx}
\usepackage{mathtools}
\usepackage{empheq}
\usepackage{siunitx}
\usepackage{wrapfig}
\usepackage{cancel}
\usepackage{url}
\usepackage{rotating} 
\usepackage{makecell}
\usepackage{multirow}
\usepackage{cite}
\usepackage{siunitx}
\usepackage[normalem]{ulem}
\newif\ifrevision
\revisionfalse 

\ifrevision
    \newcommand{\delete}[1]{\deletioncolor{\sout{#1}}}
    \newcommand{\add}[1]{\additioncolor{#1}}
\else
    \newcommand{\delete}[1]{}
    \newcommand{\add}[1]{#1}
\fi

\newcommand{\Tref}[1]{Table~\ref{#1}}
\newcommand{\eref}[1]{Eq.~\eqref{#1}}

\newcommand{\fref}[1]{Fig.~\ref{#1}}
\newcommand{\Fref}[1]{Figure~\ref{#1}}
\newcommand{\sref}[1]{Sec.~\ref{#1}}

\usepackage{xcolor}
\newcounter{todos}
\AtEndDocument{\ifnum\value{todos}>0 \PackageWarning{TODOS}{There are \arabic{todos} todos left in this paper! Fix them before submitting the paper!} \fi}

\newcommand{\deletioncolor}{\textcolor[rgb]{0.6,0,0}}

\newcommand{\argmin}{\mathop{\mathrm{argmin}}\limits}
\newcommand{\rank}[1]{\mathsf{rank}\left(#1\right)}		
\newcommand{\norm}[1]{\left\|#1\right\|}				
\newcommand{\doubleR}{\mathbb{R}}

\newcommand{\V}[1]{\ensuremath{\mathbf{#1}}}

\usepackage{xspace}
\makeatletter
\DeclareRobustCommand\onedot{\futurelet\@let@token\@onedot}
\def\@onedot{\ifx\@let@token.\else.\null\fi\xspace}
 
\def\ie{\emph{i.e}\onedot}

\def\wrt{w.r.t\onedot} 
\def\etal{\emph{et al}\onedot}

\makeatletter
\newcommand{\figcaption}[1]{\def\@captype{figure}\caption{#1}}
\newcommand{\tblcaption}[1]{\def\@captype{table}\caption{#1}}
\makeatother

\newcommand{\normalize}[1]{\eta\left(#1\right)}
\input{macro}

%% file: macro.tex
\newcommand{\journaltitle}{DP-SfM: Dual-Pixel Structure-from-Motion \\without Scale Ambiguity}
  {\color{blue}}
  {}

%% file: sec/0_abstract.tex
\begin{abstract}
Multi-view 3D reconstruction, namely, structure-from-motion followed by multi-view stereo, is a fundamental component of 3D computer vision. In general, multi-view 3D reconstruction suffers from an unknown scale ambiguity unless a reference object of known size is present in the scene. In this article, we show that multi-view images captured using a dual-pixel (DP) sensor can automatically resolve the scale ambiguity, without requiring a reference object or prior calibration. Specifically, the defocus blur observed in DP images provides sufficient information to determine the absolute scale when paired with depth maps (up to scale) recovered from multi-view 3D reconstruction. Based on this observation, we develop a simple yet effective linear method to estimate the absolute scale, followed by the intensity-based optimization stage that aligns the left and right DP images by shifting them back toward each other using cross-view blur kernels.
Experiments demonstrate the effectiveness of the proposed approach across diverse scenes captured with different cameras and lenses.
\delete{Our code and data will be released upon acceptance.}
\add{Code and data are available at https://github.com/lilika-makabe/dp-sfm-tpami.git}
\end{abstract}

%% file: sec/01_intro.tex
\section{Introduction}
\label{sec:intro}
Multi-view 3D reconstruction is a fundamental problem in 3D computer vision. From multi-view images of a scene taken from unknown viewpoints, structure-from-motion~(SfM) estimates camera poses and sparse 3D tie points of the scene. 
Subsequently, multi-view stereo~(MVS) uses the estimated camera poses to reconstruct the dense shape of the scene. 
\add{
However, since perspective projection is invariant to a global scaling of the 3D scene, the geometry and camera poses estimated by SfM, and accordingly the dense shape reconstructed by MVS, can only be determined up to an arbitrary global scale~\cite{Hartley2004}.
}
It is known that the scale ambiguity in SfM is inherent even with known camera intrinsics and cannot be resolved without external information. 
\delete{In this setting, the subsequent MVS can only recover dense shapes up to scale.}
A classical yet most reliable approach to resolving the ambiguity is to place a reference object of known size, such as a calibration target, in the scene to determine the absolute scale. 
However, placing such a reference object in the scene is not always feasible, so relaxing this requirement is desirable for diverse applications.

In this paper, we propose a method for automatically determining the absolute scale in multi-view 3D reconstruction using dual-pixel (DP) cameras whose intrinsics are calibrated. 
\begin{figure}[t]
    \centering
    \includegraphics[width=\linewidth]{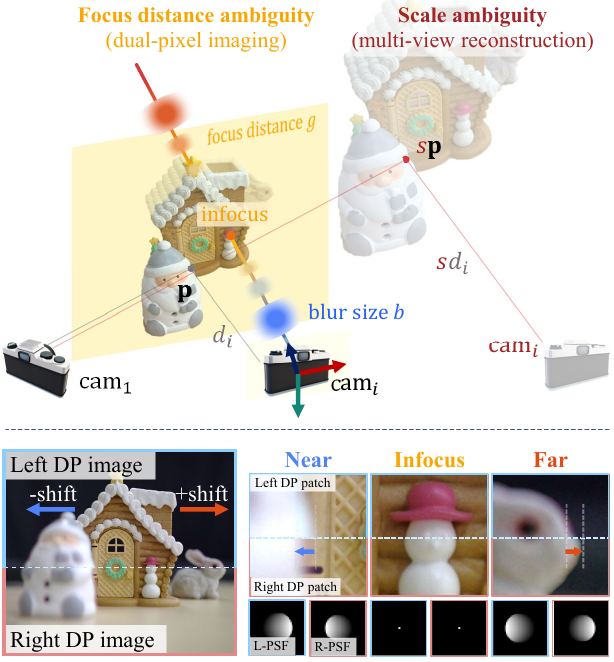}
    \caption{
    Top: Our method determines the absolute scene scale of SfM and MVS from multi-view DP images.
    We put together the scale ambiguity in SfM and MVS and the focus distance ambiguity of DP images. Our method jointly resolves the ambiguities. 
    Bottom: Visualization of DP disparity and corresponding blur, \ie, PSF kernels. 
The two sub-aperture views yield opposite disparity directions for near and far points.
    }
    \label{fig:teaser}
\end{figure}
DP sensors are designed to assist the camera's auto-focus system and are becoming popular since they are equipped with Canon EOS cameras and Google Pixel smartphones.
Each pixel on a DP sensor has two photodiodes to acquire two sub-aperture views. These views enable the estimation of \emph{defocus blurs} in regions away from the focal plane, although the blur sizes depend on the unknown focus distance. We show that, by relating the DP defocus blur sizes with unknown focus distances to the dense depth maps up to scale obtained from SfM and MVS, the absolute scale of the scene can be fully determined without any external information.

The key idea of our method is coupling two ambiguous measurements for resolving the ambiguities: One is the linear {\em scale ambiguity} of depth values computed from SfM and MVS, and the other is the {\em focus distance ambiguity} that affinely relates the reciprocal of depths with DP defocus blurs. 
As illustrated in \fref{fig:teaser}, the size of defocus blur $b$ is related to unknown focus distance $g$
and unknown scene scale $s$, with which scene point $\V{p}$ may be scaled to $s\V{p}$ retaining its projections {onto each camera view}.
By putting them together, our method formulates a regression problem to jointly estimate the absolute scene scale and the per-view focus distances. 
We exploit the fact that DP's two sub-aperture views faithfully yield defocus blur estimates, significantly more reliably than just a single view from an ordinary camera. 

The preliminary version of this work appeared in \cite{ashida2024}. Although the previous work produces reasonable results, it heavily relies on the accuracy of blur sizes that are independently estimated from image patches~\cite{DefocusDisparity}. As a result, the final scale estimate is susceptible to the blur size estimation.
In this extended version, we improve the scale estimation by introducing the intensity-based optimization stage, leveraging cross-view DP image alignment via re-blurring with point spread functions (PSFs) parameterized by the estimated scale and focus distances.
Furthermore, we demonstrate that using DP images captured with multiple apertures significantly improves accuracy over the state-of-the-art methods.

\vspace{2mm}
\noindent \textbf{Contributions.}
The chief contribution of this paper is to propose a method to determine the absolute scale in uncalibrated multi-view 3D reconstruction using DP images. To this end, we combine a linear solution method for simultaneously determining the global scene scale and per-view focus distances and the intensity-based optimization stage using a discretized search that minimizes cross-view DP image differences.

%% file: sec/02_related.tex
\section{Related Work}
\label{sec:related}

\paragraph{Scale estimation in SfM}

Since SfM alone cannot recover the absolute scale from the image information, it is necessary to introduce an additional cue that provides absolute scale information. A simple and reliable approach is to place a reference object of known size, such as markers and ground control points (GCPs), which is widely employed in commercial photogrammetry software. 
Domain-specific cues are often used to avoid manual placement of reference objects, for example, urban objects~\cite{sucar2018bayesian,frost2018recovering}, ground surfaces~\cite{SfM_autonomus_ground, zhou2016reliable, zhou2019ground}, and human faces~\cite{knorr2016leveraging}.
To implicitly exploit the scale information of scene objects, recent learning-based methods use monocular depth estimation providing absolute scale~\cite{roussel2019monocular,rukhovich2019estimation, zoedepth, depth_anything_v1, depth_anything_v2}.

Since it is not always possible to assume reference objects in target scenes, external sensors are often used to recover the scale information.
The use of stereo cameras with known baselines~\cite{engel2015large,SfM_modal} or using LiDAR~\cite{giubilato2018scale} are typical examples, while inertial sensors~\cite{nutzi2011fusion,zhang2022towards} and odometry information~\cite{SfM_nonholonomic,lee2018stability} are also widely used in robotics applications. 
Placing a refraction plate in front of the camera provides scale information using multiple capture~\cite{SfM_refraction1, SfM_refraction2}.

Similar to our work, defocus has been used to determine the absolute scale in prior studies.
{\em Calibrated} depth-from-defocus is used with SfM-based depths to recover the absolute scale by assuming a known and fixed focus distance~\cite{wohler2009monocular,shiozaki2017eliminating}. 
A recent study~\cite{toshiba} relaxes the assumption by jointly estimating the physical focal length and scene scale.
Although their use of the thin-lens model is similar to ours, their method requires pre-training or calibration for a learning-based blur size estimation~\cite{toshiba_amap} that depends on the hardware setup and focal lengths. 
In contrast, our study uses {\em uncalibrated} defocus blurs, treating the focus distance as unknown. The uncalibrated defocus blurs are reliably obtained from DP images, 
and thereby, our method recovers the absolute scales without prior knowledge, training data, or additional equipment.

\paragraph{DP imaging and applications}
Unlike conventional sensors, a single pixel in a DP sensor is divided into two photodiodes, one on the left and one on the right, which produces disparity.
Beyond its primary use of fast auto-focus~\cite{RevisitAF,LearnAF} implemented in several commercial cameras, applications of DP imaging are actively studied. 
DP's disparity gives effective cues to estimate the Point Spread Functions (PSFs) of defocus blurs~\cite{DefocusDisparity}, whose direct application is defocus deblurring~\cite{DeblurringDP,Pan_2021_CVPR,Abuolaim_2021_CVPR,Abuolaim_2021_ICCV,Xin_2021_ICCV,Yang_2023_CVPR,WACV2022,ICTC2021} and blur synthesis~\cite{Synthetic}.
The defocus blur information from DPs improves the performance of monocular depth estimation~\cite{LearningDP,DefocusDisparity,Kim_2023_CVPR}. For this application, a DP image dataset with the ground truth depths has been developed~\cite{li2023learning}.
The DP-based depth has been used for domain-specific applications, such as reflection removal~\cite{ReflectionRemove} and facial shape recovery~\cite{kang2021facial}.
A recent method combines DP imaging and traditional two-view stereo for depth estimation~\cite{Du2Net}.
Taking advantage of DP's effectiveness in acquiring defocus blur information, our work attempts to determine the scene scale automatically in multi-view 3D reconstruction.

\section{Preliminary: DP Imaging and Defocus Blur}
Unlike conventional sensors, DP sensors capture two sub-aperture images by the left and right photodiodes in a single shot.
The difference between the two images is caused by the difference in the PSFs of the left and right DP views~\cite{DefocusDisparity}. 
\Fref{fig:teaser}~(Bottom) shows the visualization of the difference between the two DP views and the corresponding PSFs.
There is no difference between the two views if the depth of a scene point is in focus.
When the scene point is out of focus, \ie, located further away or closer to the camera from the focal plane, the disparities between the two views appear in opposite directions. 
Unlike traditional two-view stereo, the disparity is caused by defocus blurs generated by the PSFs for each view.
It is known that the signed defocus blur size \hbox{$b \in \mathbb{R}$} is proportional to the \emph{defocus} disparity $d$ between two views~\cite{LearningDP}.

DP imaging makes the estimation of defocus \mbox{blur size $b$} tractable~\cite{DefocusDisparity} without external information or learning-based methods. 
Since the DP PSFs $(\mathbf{H}_l,\mathbf{H}_r)$ can be accurately modeled as a parametric function of \mbox{the blur size $b$}, the per-patch blur size is estimated from the DP image patches $(\mathbf{G}_l,  \mathbf{G}_r)$ by minimizing the cross-view loss:  
\begin{equation}
\label{eq:loss_modeling}
    b = \argmin_b \| \mathbf{G}_l * \mathbf{H}_r(b) - \mathbf{G}_r * \mathbf{H}_l(b) \|_F,
\end{equation}
assuming the depth is constant within the patch, where $*$ is a convolution operator \add{and $\norm{\cdot}_F$ is the Frobenius norm}.
The loss measures the difference between the left DP patch blurred using the right-view PSF and the right DP patch blurred using the left-view PSF.
Intuitively, this operation shifts each DP image toward the opposite sub-aperture view, effectively aligning the two image patches. This cross-view loss enables the accurate estimation of the underlying defocus blur size.

%% file: sec/03_method.tex
\begin{figure*}[t]
    \centering
    \includegraphics[width=\linewidth]{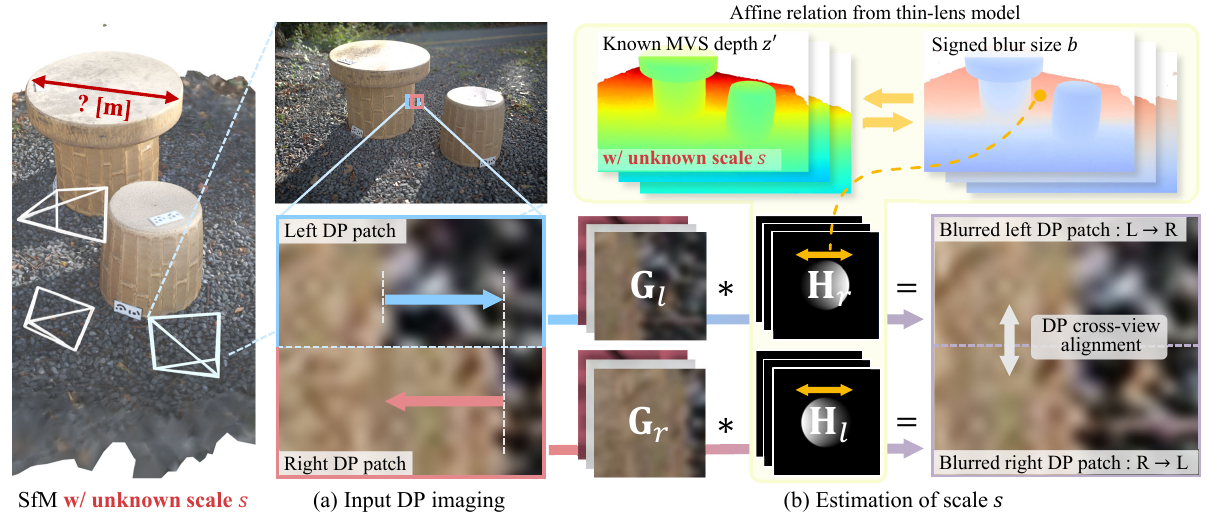}
    \caption{Overview of our method that provides the absolute scale to multi-view 3D reconstruction (SfM and MVS) just by capturing the scene using DP cameras. (a) The disparity between DP views provides the cue to estimate the defocus blur size. (b) From the affine relation between the (inverse) depth and the defocus blur size, our method simultaneously estimates the absolute scene scale and focus distances. }
    \label{fig:flow}
\end{figure*}

\section{Proposed Method} \label{sec:method}
The overview of our method is illustrated in \fref{fig:flow}. Our method infers the unknown scene scale in multi-view 3D reconstruction just by capturing the images with a DP camera.
DP images provide a useful cue for blur size estimation, as described above, without requiring any external information. 
We show in \fref{fig:flow} that the estimated blur size and (inverse) depth from SfM and MVS have an affine relationship, and thus, the unknown scene scale and focus distance can be estimated simultaneously by linear regression.

We begin by describing the ambiguities that arise in our setting, then describe the solution method for resolving them.

\paragraph{Assumptions}
We assume that camera intrinsics, such as focal lengths and aperture sizes (f-numbers), are known as they are usually accessible from the camera system. 
Although defocus blurs slightly change the center of projection, we assume the change is sufficiently small and ignore their effect on the intrinsics.
As stated, we use DP images taken from multiple viewpoints as input.
Our method does not assume pre-calibrated viewpoints nor reference objects in a scene. 

\subsection{Scale and Focus Distance Ambiguities}
There are two ambiguities in our setting: one is the scale ambiguity in SfM and MVS, and the other is the focus distance ambiguity in DP images.

\paragraph{Scale ambiguity in SfM and MVS}
Camera positions and 3D point locations recovered by SfM and the subsequent MVS have a scale ambiguity even with known camera intrinsics. Let us consider a 3D point $\V{p}= [x, y, z]^\top$ projected to the corresponding 2D point $\V{u} = [u, v, 1]^\top$ in the homogeneous coordinates with a projection matrix $[\V{R} | \V{t}]$ formed by a rotation matrix $\V{R} \in SO(3)$ and a translation vector $\V{t} \in \mathbb{R}^3$. With known camera intrinsics, we can set the intrinsics to be identity and write the 3D-to-2D perspective projection in some Euclidean world coordinate system as
\begin{equation}
\V{u} \sim  \begin{bmatrix}\V{R} & \V{t} \end{bmatrix} \begin{bmatrix}\V{p} \\ 1\end{bmatrix}
=\frac{1}{s} \begin{bmatrix}\V{R} & s \V{t} \end{bmatrix} \begin{bmatrix} s \V{p} \\ 1\end{bmatrix}, 
\end{equation}
where $s \in \mathbb{R}$ is an unknown scaling factor.
As such, the depth $z'$ obtained by SfM and MVS is linearly related to the true depth $z$ by the unknown scaling factor $s$ as
\begin{eqnarray}
\label{eq:depth_ambiguity}
z = s z'.
\end{eqnarray}
This is the inherent ambiguity in SfM, which we are going to resolve in this paper. 


\paragraph{Focus distance ambiguity in DP images}
As described by Wadhawa~\etal~\cite{Synthetic} and Garg~\etal~\cite{LearningDP}, the signed defocus blur size \hbox{$b \in \mathbb{R}$} is related to scene depth $z$, focal length $f$, lens aperture diameter $l$, and the focus distance $g$ in $\mathbb{R}$ as
\begin{eqnarray}
\label{eq:focas_distance_ambiguity}
b \approx \frac{l f}{1-f/g}\left(\frac{1}{g} - \frac{1}{z}\right)
\end{eqnarray}
using the paraxial and thin-lens approximations.
The blur size $b$ in DP images varies with the focus distance $g$ even when the same scene depth $z$ is recorded with the same camera parameters $(l, f)$. Therefore, even with the blur size $b$ from DP images, its relationship to depth $z$ is ambiguous due to unknown focus distance $g$ and depth $z$ as
\begin{eqnarray}
b \approx \frac{l f}{1-f/g}\left(\frac{1}{g} - \frac{1}{z}\right) = \frac{l f}{1-f/g'}\left(\frac{1}{g'} - \frac{1}{z'}\right), 
\end{eqnarray}
\ie, there exists a set of pairs like $(g, z)$ and $(g', z')$ that yields the same blur size $b$.
The blur size $b$ is signed, \ie, it becomes positive when the scene point is farther away from the focal plane and negative when nearer to the camera.

\subsection{{Initial Estimation}}
Given the patch-wise blur sizes $b_{ij} \in \doubleR$ obtained from DP images, patch-wise depths $z'_{ij} \in \doubleR_+$ from SfM and MVS in the $i$-th view at {$j$-th patch}, and known per-view focal lengths $f_i$ and lens aperture $l_i$, our goal is to determine unknown scene scale $s \in \mathbb{R}_+$ and per-view focus distances $g_i \in \doubleR_+$.
Since we use per-view intrinsics, our method can be used in a practical setting of multi-view 3D reconstruction, in which camera intrinsics may vary across viewpoints.

We put together the two ambiguous observations, \ie, depths up to scale and blur sizes with unknown focus distances, by substituting \eref{eq:depth_ambiguity} to \eref{eq:focas_distance_ambiguity} as 
\begin{align}
\label{eq:solution01}
    b_{ij} &= \frac{l_{i}f_{i}}{1-f_{i}/g_{i}}\left( \frac{1}{g_{i}} - \frac{1}{sz'_{ij}} \right).
\end{align}
Denoting $\bar{g}_i = \frac{1}{g_i}$ and $\bar{s} = \frac{1}{s}$ as $g_i$ and $s$ are both non-zero, 
\eref{eq:solution01} can be rewritten as
\begin{align}
    \frac{b_{ij}z'_{ij}}{f_i} &= \underbrace{z'_{ij} \left( b_{ij} + l_{i} \right)}_{\gamma_{ij}} \bar{g_{i}} - l_{i}\bar{s} = \gamma_{ij} \bar{g_{i}} - l_{i}\bar{s}, \label{eq:one_pixel}
\end{align}
which is linear \wrt the unknown parameters $\bar{g_i}$ and $\bar{s}$.

From observations of multiple views and patches, we can construct an over-determined system of linear equations from \eref{eq:one_pixel}. Given the number of views $n$ and the number of valid patches $p_i$ in the $i$-th view, we have $\sum_{i=1}^{n} p_i$ equations, which can be written in a matrix form as:
\begin{align}
{\tiny
\underbrace{
    \begin{bmatrix}
    \gamma_{11}       &                &           &        & -l_1\\
    \vdots            & \multicolumn{3}{c}{\V{0}}                              &  \vdots\\
    \gamma_{1 p_1}    &                 &           &        & -l_1\\
                      & \gamma_{21}     &           &        & -l_2\\
    \V{0}             & \vdots          & \multicolumn{2}{c}{\V{0}}       &\vdots\\
                      & \gamma_{2 p_2}  &          &         & -l_2\\
                      &                &   \ddots &        & \vdots\\      
                      &                &           & \gamma_{n1} & -l_n \\
    \multicolumn{3}{c}{\V{0}}                      & \vdots  & \vdots\\     
                      &                &           & \gamma_{n p_n}  & -l_n
    \end{bmatrix}}_{\mathbf{A}}
\underbrace{
    \begin{bmatrix}
    \bar{g}_1 \\
    \bar{g}_2 \\
    \vdots \\
    \bar{g}_{n} \\
    \bar{s}
    \end{bmatrix}}_{\mathbf{x}} 
    =
\underbrace{
    \begin{bmatrix}
    b_{11}  z'_{11} / f_1 \\
    \vdots \\
    b_{1 p_{1}}  z'_{1 p_{1}} / f_1 \\
    b_{21}  z'_{21} / f_2 \\
    \vdots \\
    b_{2 p_{2}}  z'_{2 p_{2}} / f_2 \\
    \vdots \\
    b_{n 1}  z'_{n 1} / f_n\\
    \vdots \\
    b_{n p_{n}}  z'_{n p_{n}} / f_n
    \end{bmatrix}}_\mathbf{b}. 
\label{eq:linear_system} 
}
\end{align}
Since we have $n+1$ unknowns in vector $\V{x}$, given $\V{b} \neq \V{0}$, the condition for obtaining a unique (approximate) solution is $\rank{\V{A}} = n+1$, which is, in practice, easy to achieve because a large number of patches are available. The minimum setting to determine scale $s$ is selecting two patches with distinct depths in one of the multi-view images, which makes a full-rank $2\times 2$ matrix $\V{A}$ for two unknowns $\V{x} = [\bar{g}_i, \bar{s}]^\top$. 

We can select many patches from multiple views to rely on robust estimation. With a sufficiently large number of selected patches, we construct a tall and skinny matrix $\V{A}$ and use $\ell_1$ residual minimization to effectively disregard outliers \add{due to unmodeled errors in the thin-lens assumption, such as asymmetric PSFs or optical aberrations~\cite{toshiba_amap, Abuolaim_2021_ICCV}, } as 
\begin{align}
    \V{x}_\mathrm{ini} = \argmin_{\mathbf{x}} \left\|\mathbf{A} \mathbf{x} - \mathbf{b}\right\|_1. 
\end{align}
We use the Iteratively Reweighted Least Squares~(IRLS) method~\cite{IRLS} for deriving the approximate solution $\V{x}_\mathrm{ini}$. Obviously, other choices of robust estimation methods, such as RANdom SAmple Consensus (RANSAC)~\cite{RANSAC}, are viable for alleviating the impact of outliers.

\subsection{View Selection}
So far, we have assumed that the blur sizes $b_{ij}$ in initial estimation are available. For obtaining the blur sizes, we use the method proposed by Punnappurath~\etal~\cite{DefocusDisparity}. 
For our case, we only require blur sizes for a small number of samples but not for all the patches since our goal is determining the scene scale $s$ using \eref{eq:linear_system}. 
Therefore, we choose a small subset of patches and views that have reliable blur size estimates. 

Since the scale $s$ can be estimated for each view as long as the depths up to scale are available, we first solve~\eref{eq:linear_system} for each view independently using the selected patches. Ideally, the scales estimated for each view should match; however, in practice, the estimates vary due to outliers. 

We exclude views with insufficient blur size variation, where the difference between the maximum and minimum blur sizes is less than or equal to $T_p$ pixels. Such a view is not preferable since it indicates little depth variation between the patches, which makes the linear system of \eref{eq:linear_system} unstable. We also exclude views with negative scale estimates, which
may result from motion blur or inaccurate blur or depth estimation.

Finally, from the remaining views, we select $N_{v}$
views that are close to the median of the estimated scales from all views. 
Once the appropriate views are selected, by using all the selected patches within the views, we create the linear system of~\eref{eq:linear_system} and obtain the re-estimated scale $s^*$ and the scale unknown focus distances $g_i'^*\triangleq{g_i^*/s^*}$.
Patch selection strategy can be found in the supplementary material.

\begin{figure}[t]
    \centering
    \includegraphics[width=\linewidth]{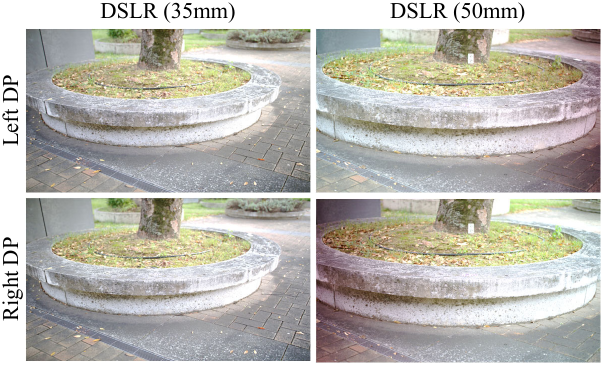} 
    \caption{Example of DP images captured with different lenses.
The right column shows spatially varying, channel-dependent artifacts, where the image noticeably shifts toward red on one side. These artifacts are not present in the DP image captured with the other lens, shown in the left column.
    }
        \vspace{1mm}
    \label{fig:dpimage_ex}
\end{figure}
\subsection{\add{Intensity-based Optimization}}
The blur size computed using the initial depth scale $s^*$ does not always match the actual image observations. In particular, the blur size from the initial guess may fail to correctly shift the dual-pixel (DP) image patch in the opposite direction and align the views, unlike the ideal case illustrated in \fref{fig:flow}. This is mainly because the initial global scale is obtained using blur sizes estimated independently from individual patches, without explicitly using the thin-lens model described in \eref{eq:focas_distance_ambiguity}. Therefore, in the final step, we refine the scale estimation by minimizing a cross-view alignment loss. This refinement uses the same patches and views selected at the initial estimation but incorporates the thin-lens model explicitly.

Specifically, we fix the scale-ambiguous focus distance $g_i'^*$ from the initial guess and optimize only the scale $s$. For each pair of the left and right DP image patches, $\mathbf{G}^{(l,c)}_{j}\in\doubleR^{m\times m}$ and $\mathbf{G}^{(r,c)}_{j}\in\doubleR^{m\times m}$, at $j$-th patch in the $i$-th view and channel $c\in\{R,G,B\}$, we minimize the following objective:
\begin{equation}
\label{eq:loss_refine}
\footnotesize
    \mathcal{L}
    = \sum_{i, j, c} 
    \left\| \normalize{\mathbf{G}^{(l,c)}_{j} * \mathbf{H}_{{i,j}}(s)} - \normalize{\mathbf{G}^{(r,c)}_{j} * \mathrm{flip}\left({\mathbf{H}_{{i,j}}(s)}\right)} \right\|_{1,1},
\end{equation}
similar to \cite{DefocusDisparity}.
Here, $m$ is the patch size and the operator $\mathrm{flip}()$ performs horizontal flipping, which is used to obtain the left-view PSF from the right-view PSF $\V{H}_{ij}$, preserving the mirror symmetry of the left/right DP PSFs as illustrated in \fref{fig:teaser}~(Bottom).
The norm $\|\cdot\|_{1,1}$ denotes the element-wise $\ell_1$ norm over the image patch, and the operation $\normalize{\cdot}$ represents per-patch normalization, defined as $\normalize{\V{X}}=\V{X}/\norm{\V{X}}_{1,1}$. 
For the parameterization of \mbox{right PSF $\V{H}_{ij}$}, we use the same model as in \cite{DefocusDisparity}, defined as:
\begin{equation}
    \begin{aligned}
    \mathbf{H}_{ij}(s) &= \sum_{k=0}^{|2r_{ij}|} \left\{ \mathsf{C}(0,0; r_{ij}) \odot \mathsf{C}\left( k\frac{r_{ij}}{|r_{ij}|}, 0 ; r_{ij}\right) \right\} \\
    r_{ij}(s) &= \frac{p}{2} b_{ij} = \frac{p}{2} \cdot \frac{l_{i} f_{i}}{1 - f_{i} / sg'^*_{i}} \left( \frac{1}{sg'^*_{i}} - \frac{1}{s z'_{ij}} \right),
    \end{aligned}    
\end{equation}
where $r_{ij}$ is the signed blur radius in the pixel coordinates, $p\in\doubleR$ is the sensor pitch, and the blur size $b_{ij}$ depends on the unknown depth scale $s$, as defined in \eref{eq:solution01}. The operators $|\cdot|$ and $\odot$ denote the absolute value and the element-wise multiplication, respectively. $\mathsf{C}(x, y; r)$ represents a circular disk of radius $|r|$ centered at $(x, y)$.

We found that the normalization operation $\eta$ is essential, as decoded DP images captured with certain lenses exhibit spatially varying artifacts, as shown in \fref{fig:dpimage_ex}. 
To mitigate their impact during the intensity-based optimization stage, we incorporate per-patch normalization directly into the objective function.

Finally, we search for the optimal global scale $s_\mathrm{optim}$ over a candidate set $\mathcal{S}$, which minimizes the loss as 
\begin{equation}
    s_\mathrm{optim} = \argmin_{s \in \mathcal{S}} \mathcal{L}.
\end{equation}
The candidate set $\mathcal{S}$ is uniformly sampled around the initial global scale $s^*$ within a range of $100 \pm T_s\%$. Empirically, 100 candidates suffice; for $T_s=80\%$, this yields a resolution of $1.6\%$ of the initial global scale $s^*$, corresponding to at most a few centimeters in our scenes.
We found that optimizing both the global scale $s$ and focus distances $g_i$ from random initialization using the cross-view alignment loss is highly unstable due to the problem's high flexibility.
This highlights the importance of a good initial estimate and motivates the use of discrete search in the final stage.
\add{Further implementation details can be found in the supplementary material.}

%% file: sec/04_experiments.tex
\section{Experiment}
 \begin{figure*}[tp]
\centering
    \includegraphics[width=1.\linewidth]{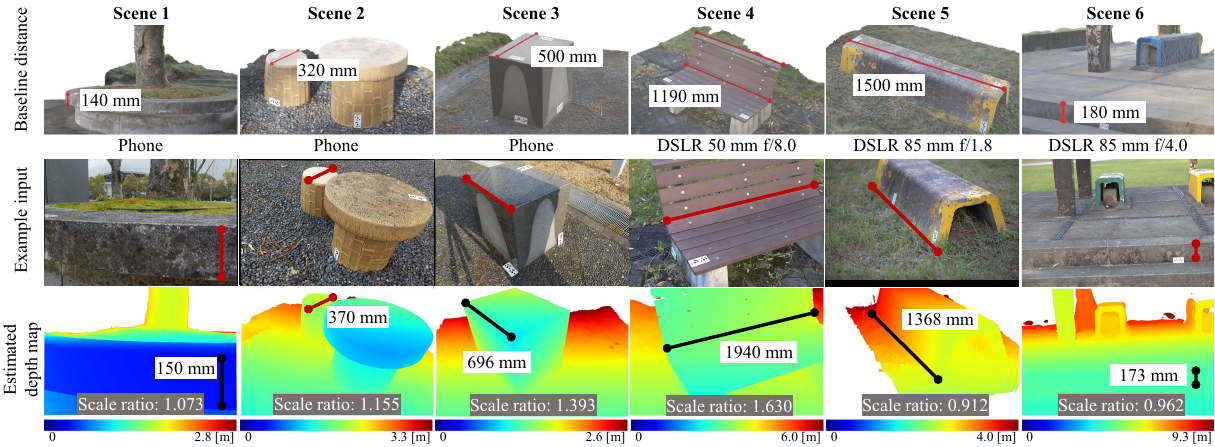}  
    \caption{
    Visual results of real scenes captured by the DSLR (Canon EOS 5D Mark IV) and Phone (Pixel~4). Our method yields reasonable estimates of scene scales just by using a DP camera for multi-view capture, which is theoretically unrecoverable by standard SfM alone. 
    In the first row, we show the absolute baseline length (red-annotated line), obtained from the ground-truth 3D model.
    For each scene and camera setting, we show the estimated absolute distance after scale recovery. For Pixel 4 results, we omit the camera setting notation.
    }
    \label{fig:vis_results}
\end{figure*}       

 \begin{figure}[tp]
\centering
    \includegraphics[width=1.\linewidth]{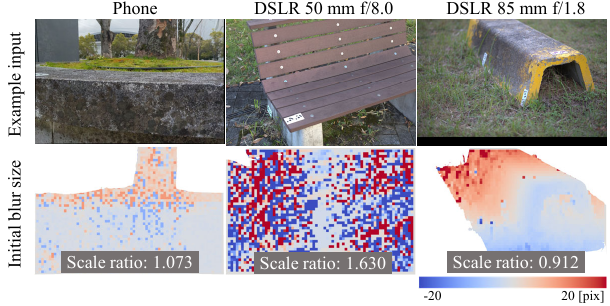}  
    \vspace{-6mm}
    \caption{
    Visual results of the estimated blur sizes using the method from \cite{DefocusDisparity}, which are used to compute the initial global scale and the scaled focus distances.
    }
    \label{fig:initial_blur}
\end{figure}       
\label{sec:experiment}
We evaluate our scale estimation method both quantitatively and qualitatively.
We use multi-view DP images captured in six scenes
with different capturing settings, such as, cameras (DSLR and smartphone), lenses, and aperture sizes. 

\paragraph{Recording setting}
We use two different types of cameras equipped with a DP sensor: a DSLR camera and a smartphone.
For the DSLR camera setting, we use Canon EOS 5D Mark IV with three different lenses: EF35mm F1.4L II USM, EF50mm F1.8 STM, and EF85mm F1.8 USM. To assess the effect of aperture size, we vary the aperture per scene, using the lens's maximum aperture ($f/1.4$ or $f/1.8$), $f/4.0$, and $f/8.0$, depending on the scene. The images from Canon's DLSR camera are recorded with CR2 format, and the left and right DP views are extracted using DPRSplit\footnote{DPRSplit~\url{https://www.fastrawviewer.com/DPRSplit}, last accessed on June 15, 2025.}.
For a more accessible setup, we also use a Google Pixel~4 smartphone, which has a focal length of $4.38$~mm and a fixed aperture of $f/1.73$. 
The left and right DP views are obtained using the capturing software used in~\cite{GargDualPixelsICCV2019}.


For each scene for each camera setting, we capture about $30$~images using a DSLR, and $60$~images using a smartphone.
%
For fair comparisons across different lenses and aperture sizes, for each viewpoint, we use a tripod to keep the camera fixed and record images with different camera settings. 
To evaluate the scale estimates, we obtain the ground-truth scene scale by placing markers with known sizes in the scenes.

\paragraph{Evaluation metric}

As a quantitative evaluation metric, we use the scale ratio $r_{s}$ of the estimated scene scale $s_{\rm est}$ to the ground-truth scale $s_{\rm gt}$, which is defined as follows:
\[
\text{Scale ratio~~} r_{s} = \frac{s_{\rm est}}{s_{\rm gt}}.
\]
In the above metric, closer to $1$ indicates good scale estimation. The scale ratio smaller than $1$ indicates that the scale is under-estimated, \ie, the estimated scale is smaller than the true scale, and the greater scale ratio corresponds to the over-estimation. 
While the scale ratio $r_{s}$ is defined for each scene and camera setting, we also define the average error of the scale ratios $e_s$ to characterize the overall effectiveness as
\[
\text{Average error of scale ratios~~} e_s = \frac{1}{n_s} \sum \left|1-r_s \right|,
\]
where $n_s$ is the number of settings.

\begin{table}[t]
    \centering
    \resizebox{1.0\linewidth}{!}{
    \begin{tabular}{
        @{\hspace{0.25em}}l|c@{\hspace{0.25em}}c@{\hspace{0.25em}}c|c@{\hspace{0.25em}}c@{\hspace{0.25em}}c|c@{\hspace{0.25em}}c@{\hspace{0.25em}}c|c|c
    }
    \toprule
    & \multicolumn{3}{c|}{DSLR (35mm)} 
    & \multicolumn{3}{c|}{DSLR (50mm)} 
    & \multicolumn{3}{c|}{DSLR (85mm)} 
    & Phone 
    & \multirow{2}{*}{$e_s$} \\
    
    & $f/1.4$ & $f/4.0$ & $f/8.0$ 
    & $f/1.8$ & $f/4.0$ & $f/8.0$ 
    & $f/1.8$ & $f/4.0$ & $f/8.0$ 
    & $f/1.73$ & \\
    \midrule


\text{Scene1} & 1.188 & 0.941 & -- & 1.152 & 1.014 & -- & 1.173 & 1.095 & -- & 0.979 & \textbf{0.100} \\
\text{Scene2} & 1.284 & 1.129 & -- & 1.451 & 1.188 & -- & 1.236 & 1.164 & -- & 1.155 & \textbf{0.230} \\
\text{Scene3} & 1.141 & 1.116 & -- & 1.292 & 1.500 & -- & 1.074 & 1.035 & -- & 1.393 & \textbf{0.222} \\
\text{Scene4} & 1.121 & 0.855 & 0.958 & 1.044 & 1.000 & 1.630 & 0.998 & 1.038 & 1.073 & 1.397 & \textbf{0.149} \\
\text{Scene5} & 1.087 & 0.818 & 0.734 & 1.069 & 1.041 & 1.137 & 0.912 & 0.940 & 1.112 & 1.418 & \textbf{0.146} \\
\text{Scene6} & 1.063 & 0.888 & -- & 0.998 & 0.983 & -- & 1.003 & 0.962 & -- & 1.320 & \textbf{0.079} \\
\midrule
$e_s$  & \textbf{0.147} & \textbf{0.124} & \textbf{0.154} & \textbf{0.168} & \textbf{0.127} & 0.383 & \textbf{0.096} & \textbf{0.072} & \textbf{0.093} & \textbf{0.284} & \textbf{0.153} \\
    
    \bottomrule
    \end{tabular}
    }
    \caption{Scale ratio $r_{s}$ between the ground truth and the estimated scale for each scene, lens, and aperture size. Maximum aperture varies depending on the lens (\ie, $f/1.4$ or $f/1.8$). $e_s$ represents the average errors of the corresponding row/column/total. ``-'' indicates that we did not record the scene under the setting. Values with smaller average errors than the initial estimates are shown in \textbf{bold}. }
     \label{table:per-setting-results}
\end{table}

\paragraph{Quantitative results}
The results of the scale estimation for six scenes using two DP cameras are summarized in Table~\ref{table:per-setting-results}.
The DSLR camera uses three lenses and different aperture sizes.
Overall, our method can reasonably well estimate the absolute scene scale just by using a DP camera for multi-view capture, considering the fact that scale ratio $r_s$ can range in {$[0, \infty)$}.
The average error $e_{s}$ across scenes and lenses is approximately $0.153$ as shown in the table, where the error from the initial guess is $0.266$.
From comparing the average errors for each lens, we see that the difference in lenses does not significantly affect the overall performance. 

\paragraph{Visual results}
Figure~\ref{fig:vis_results}
show visual examples of our results by the DSLR and smartphone, as well as the ground-truth depth maps.
We can confirm the same trend as in the quantitative results, where the depth maps with estimated scale well represent the actual depths. 

\begin{figure}[t]
\centering
    \includegraphics[width=\linewidth]{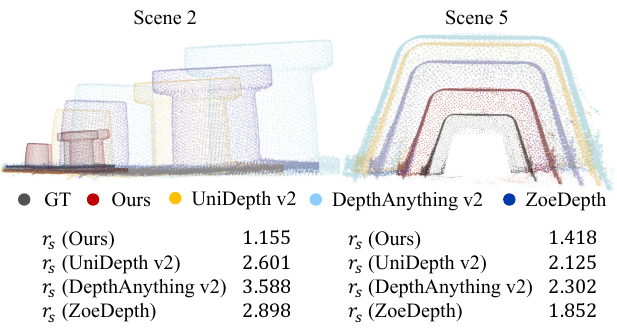}
    \caption{Comparison with \delete{monocular scale estimation by the Phone.} \add{metric depth estimation methods using Phone data.} 
    \delete{Blue and red points represent the ones of the scale ratio estimated by our method, DepthAnything~v2, and ZoeDepth, respectively.}
    \add{Each colored point cloud represents the estimated scene at an absolute scale.}
    $r_s\left(\cdot\right)$ shows the ratio between the GT and the estimated scale by each method.}
    \label{fig:comparisons_pc}
\end{figure}
\paragraph{Comparison with scale estimation by monocular depth estimation}
We compare our method with 
\add{the state-of-the-art metric depth estimation methods, UniDepth~v2\cite{piccinelli2024unidepth, piccinelli2025unidepthv2}, DepthAnything~v2\cite{depth_anything_v1, depth_anything_v2}, and ZoeDepth~\cite{zoedepth}}, for the same scenes and the lens setup we used. Since they estimate metric depth for each pixel, we compute the per-image scale as the average ratio between \delete{ZoeDepth}\add{estimated depth from each method} and MVS depth maps.
{The detailed experimental setting for \delete{monocular depth estimation}\add{comparisons} can be found in the supplementary material.}

The average error $e_s$ across scenes and lenses using UniDepth~v2, DepthAnything~v2 and ZoeDepth is \add{$0.416$, $0.519$, and $0.565$}, respectively, which is around three times larger than ours ($0.153$).
\Fref{fig:comparisons_pc} shows the visual comparison of the estimated scale using Phone data, which confirms this trend.
More detailed comparisons and further experimental results, including evaluations on additional indoor scenes, can be found in the supplementary material.

\paragraph{Multi apertures experiments}
\label{sec:multi-aperture}
As discussed in \sref{sec:method}, our method is compatible with multi-camera or multi-capture settings. One practical scenario is to capture images using a single lens but with multiple aperture sizes. To evaluate the effectiveness of leveraging such multi-setting data, we collect all images taken with different apertures but the same focal length, treat each as an individual ``view'' as in \sref{sec:method}, select the top $T_c\%$ patches from each of the $N_v$ views, and follow the exact same procedure as we did in per-camera-setting experiments in \Tref{table:per-setting-results} to estimate the global scale $s$.

\Tref{table:multi-aperture} shows the estimated global scale using both our initialization~(Initial, corresponding to~\cite{ashida2024}) and after the intensity-based optimization~(Optim) with multi-aperture images. The overall error is reduced to $9.8\%$, a significant improvement over the $15.5\%$ error observed in the per-camera-setting results shown in \Tref{table:per-setting-results}. 
The improvement is particularly notable in scenes where three different apertures are available, such as Scene~4, which previously exhibited a scale ratio $r_s$ of $1.630$ in the per-camera-setting result. This highlights the effectiveness of using multi-aperture images for accurate scale estimation.

\begin{table}[bt]
\centering
\begin{tabular}{lc@{\hskip 0.5em}c@{\hskip 0.5em}c@{\hskip 0.5em}c@{\hskip 0.5em}c@{\hskip 0.5em}c@{\hskip 0.5em}c@{\hskip 0.5em}c}
\toprule
\multirow{2}{*}{} & \multicolumn{2}{c}{\SI{35}{\milli\meter}} & \multicolumn{2}{c}{\SI{50}{\milli\meter}} & \multicolumn{2}{c}{\SI{85}{\milli\meter}} & \multicolumn{2}{c}{$e_s$} \\
\cmidrule(lr){2-3} \cmidrule(lr){4-5} \cmidrule(lr){6-7} \cmidrule(lr){8-9}
 & Initial & Optim & Initial & Optim & Initial & Optim & Initial & Optim \\
\midrule
Scene1 & 1.324 &  \textbf{1.078} & 1.090 & \textbf{0.993} & 1.209 & \textbf{1.082} & 0.208 & \textbf{0.056} \\
Scene2 & 1.286 & \textbf{1.130} & \textbf{1.232} & 1.322 & 1.297 & \textbf{1.119} & 0.272 & \textbf{0.190} \\
Scene3 & 1.229 & \textbf{1.120} & \textbf{1.351} & 1.515 & 1.160 & \textbf{1.019} & 0.247 & \textbf{0.218} \\
Scene4 & 1.207 & \textbf{0.865} & 1.311 & \textbf{1.046} & 1.256 & \textbf{1.043} & 0.258 & \textbf{0.075} \\
Scene5 & 1.198 & \textbf{1.052} & 1.163 & \textbf{1.041} & 1.224 & \textbf{1.017} & 0.195 & \textbf{0.037} \\
Scene6 & 1.184 & \textbf{0.983} & 1.169 & \textbf{1.008} & 1.131 & \textbf{0.994} & 0.161 & \textbf{0.010} \\
\midrule
$e_s$  & 0.238 & \textbf{0.089} & 0.219 & \textbf{0.156} & 0.213 & \textbf{0.048} & 0.223 & \textbf{0.098} \\
\bottomrule
\end{tabular}
\caption{Scale ratio $r_s$ between the ground truth and the estimated scale by our method, using multi-aperture images before and after the intensity-based optimization.
For each cell, the result with the smaller error is shown in \textbf{bold.}}
\label{table:multi-aperture}
\end{table}

\paragraph{Limitations and failure cases}
Although our method successfully estimates the scene scale in many cases, we find several failure modes in our experiments.
Since our method is based on defocus blur, scale estimation from pan-focus images becomes challenging. 
In particular, since the aperture of smartphone cameras is smaller than that of DSLRs, it was necessary for us to include photographs that exhibit large depth variations to produce relatively large blur, as shown in \fref{fig:initial_blur}.
Although the pan-focus images are preferable in SfM and MVS 
for better correspondence matching, in practice, small apertures cannot always be used. This is 
because the small aperture limits the amount of incoming light, requiring longer exposure times and resulting in camera motion blurs. 

The success of our method relies on the accurate estimation of both defocus blur sizes and depth maps up to scale. 
We found several examples where scale estimation is affected by errors in blur size estimation, whereas depth maps up to scale are almost always stably estimated.
The initial blur size from \cite{DefocusDisparity} in Scene~4 captured with the DSLR ($\SI{50}{\milli\meter}$ lens and $f/8.0$ aperture size) shows mostly the random blur sizes as seen in \fref{fig:initial_blur}, whose scale ratio $r_s$ reaches $1.630$. 
Most blur size estimates are far from ideal due to the target object being almost textureless or the repetition of fine-grained texture, which makes the blur size optimization more challenging. 
These failure cases can be improved by multi-aperture images as described in the previous experiment.
However, when all per-aperture results for a given camera setting show large errors, such as in Scene 3 captured with the DSLR and \SI{50}{\milli\meter} lens where the error ranges from $30\%$ to $50\%$, the error remains large even when using multi-aperture inputs due to the lack of reliable views and patches.

\add{
Beyond blur-size estimation, there also exist cases where the upstream SfM fails to produce geometrically consistent reconstructions. 
Such failures typically occur in scenes with repetitive structures, extremely low texture, or strong reflections, where SfM cannot establish stable feature correspondences. 
These cases coincide with the failure conditions of conventional SfM/MVS pipelines, and when SfM itself becomes unstable, further scale estimation is naturally difficult.
}



%% file: sec/05_conclusion.tex
\section{Discussions}
\label{sec:conclusion}
This paper has demonstrated that the scale information in multi-view 3D reconstruction, which is theoretically unsolvable by SfM and MVS themselves, can be inferred from the defocus blur information provided by DP imaging.
Our method estimates the absolute scale without relying on reference objects or data priors, but by simply capturing multi-view images with a DP camera.
The key is the unique property of DP that the reciprocal of the depth and the defocus blur sizes given by DP images are affinely related. 
Leveraging this relationship, our method resolves the global scale ambiguity in SfM by ensuring that reblurring with the estimated blur size aligns the left and right DP observations.
The experiments demonstrate that our method consistently recovers the absolute scale 
across diverse scenes and camera settings.
\section*{Acknowledgments}
This work was supported by JSPS KAKENHI Grant Numbers JP23H05491 and JP25K00142.